\documentclass[pdflatex,sn-mathphys-num]{sn-jnl}

\usepackage{graphicx}%
\usepackage{multirow}%
\usepackage{amsmath,amssymb,amsfonts}%
\usepackage{amsthm}%
\usepackage{mathrsfs}%
\usepackage[title]{appendix}%
\usepackage{xcolor}%
\usepackage{textcomp}%
\usepackage{manyfoot}%
\usepackage{booktabs}%
\usepackage{algorithm}%
\usepackage{algorithmicx}%
\usepackage{algpseudocode}%
\usepackage{listings}%
\usepackage{caption}%
\usepackage{placeins}%
\usepackage{hyperref}%
\theoremstyle{thmstyleone}%
\theoremstyle{thmstyletwo}%

\theoremstyle{thmstylethree}%

\begin{document}

\title[Article Title]{Physics-Guided Spatiotemporal Learning for Coastal Wave Peak Period Estimation from Video}


\author*[1]{\fnm{Abubakar Hamisu} \sur{Kamagata}}\email{kamagata012@gmail.com}

\author[1]{\fnm{Dharm Jat} \sur{Singh}}\email{dsingh@nust.na
}
\equalcont{These authors contributed equally to this work.}

\author[1]{\fnm{Attlee Munyaradzi} \sur{Gamundani}}\email{agamundani@nust.na}
\equalcont{These authors contributed equally to this work.}

\author[2]{\fnm{Abhishek} \sur{Srivastava}}\email{asrivastava@iiti.ac.in}
\equalcont{These authors contributed equally to this work.}

\author[1]{\fnm{Paramasivam} \sur{Saravanakumar}}\email{Paramasivam.Saravanakumar@debeersgroup.com}
\equalcont{These authors contributed equally to this work.}

\affil*[1]{\orgdiv{Computer Science}, \orgname{Namibia University of Science and Technology}, \orgaddress{\street{13 Jackson Kaujeua Street}, \city{Windhoek}, \postcode{13388}, \state{Windhoek}, \country{Namibia}}}

\affil[2]{\orgdiv{Computer Science Engineering}, \orgname{Indian Institute of Technology Indore}, \orgaddress{\street{Khandwa Road, Simrol}, \city{Indore}, \postcode{453552}, \state{Madhya Pradesh}, \country{India}}}


\abstract{Direct estimation of physically interpretable periodic signals from raw video constitutes a spatiotemporally grounded learning problem that proves to be difficult especially when facing label sparsity, lack of physical grounding and standardization benchmarks. The wave monitoring at coastal sites is one such real-world example where current deep learning approaches for estimating wave parameters using video as input suffer from physical interpretability and require some kind of intermediate data processing. In this study we propose a framework for wave peak period estimation using only video as input through three components: automated region-of-interest detection using temporal pixel variance, multi-stage Sim-to-Real transfer learning process, and physics-guided regularization of the output predictions. Various spatiotemporal architectures, including Transformer and recurrent-convolutional were compared during the stages of synthetic pretraining, silver label adaptation, and expert fine-tuning. It has been found out that LtViViT achieves the highest accuracy in its estimates, while TinyWaveNet shows superior temporal stability and oceanographic skill. Additionally, ablation studies have demonstrated that physics-guided regularization helps to follow the trends in predictions more consistently and prevent physically meaningless predictions. Moreover, Grad-CAM-based explainability analysis of the physics-guided TinyWaveNet showed that its spatial focus aligns with hydrodynamically active surf-zone regions. Overall, the findings support physics-guided, video-based deep learning as a cost-effective and operationally viable approach for long-term coastal wave monitoring, and demonstrate a transferable strategy for physically-constrained spatiotemporal regression from video under data-scarce conditions.}




\maketitle

\section{Introduction}\label{sec1}
End-to-end spatiotemporal deep learning for extracting physically meaningful periodic signals directly from raw video remains an open methodological challenge, particularly when labeled data is scarce, physical plausibility is unconstrained, and standardized benchmarks are unavailable. Recent advances in spatiotemporal modeling have used ConvLSTM and ConvGRU variants for sequence forecasting tasks \cite{abdelhady2023machine, yu2021spatiotemporal, zhang2023ocean}, while attention-enhanced transformers and CNN-xLSTM hybrids have improved performance by exploiting long-range temporal and spatial correlations \cite{liu2025regional,shi2023machine, karami2026optimizing}. Attention mechanisms and transformer architectures in particular have proven effective at capturing long-range temporal and spatial dependencies in forecasting, error correction, and video-processing tasks more broadly \cite{liu2025regional, shi2023machine, yu2021spatiotemporal}, and transformer-based video understanding has advanced rapidly in adjacent remote-sensing and detection domains \cite{jiao2023transformer}.

At the same time, physics-based and hybrid approaches have utilized domain knowledge within the network training process, resulting in increased physical accuracy and computational efficiency, consistently beating purely data-driven approaches and even traditional numerical methods on different tasks \cite{deo2025predicting, saviz2024physics, ouyang2023wave, wei2022convolutional}. Convolutional neural networks have also enabled direct regression of physical quantities from single-frame images or localized sensor data without intermediate feature engineering \cite{bai2022development, sithara2025machine, xu2024nearshore}.

Despite this progress, video-derived regression tasks remain fragmented across two largely disconnected paradigms: recurrent-convolutional models that emphasize local inductive biases for motion capture \cite{sithara2025machine, yang2020quantifying}, and transformer-based models that prioritize global context but are rarely evaluated in an end-to-end video regression setting outside non-video forecasting tasks \cite{jiao2023transformer, kaneko2025phase, yu2021spatiotemporal}.There is no standard evaluation methodology or comparative evaluation between these families of architectures for the task of direct regression from video to physical parameters, and the unavailability of publicly annotated datasets by experts further complicates any attempt at cross-evaluation of performance across different architectural designs.

Nearshore wave peak period ($T_p$), which is the period corresponding to the most energetic spectral density of the waves, is an important variable that influences wave resonance, infragravity wave production, beach evolution, and storm-induced coastal erosion \cite{gomez2024infragravity, sithara2025machine, westhuysen2012modeling, xu2024nearshore, zhou2025noncontact}. Conventional in-situ instruments such as buoys, pressure sensors, LiDAR, and radar-based remote sensing measure it reliably but are costly to deploy and maintain and offer limited spatial coverage \cite{ahmed2023low, kinsela2024nearshore, zhou2025noncontact}. Low-cost passive video monitoring has emerged as an alternative, with portable stereo systems and shore-based Video Beach Monitoring Systems (VBMS) approaching the accuracy of conventional sensors at a fraction of the cost \cite{vieira2025nearshore, venkateswarlu2025advancing}.

Classical (non-deep-learning) computer vision pipelines were the first to exploit this opportunity: Radon-transform-based wave runup detection under varying illumination \cite{almar2017wave}; modulated transfer functions and multi-taper spectral methods for timestack analysis \cite{ramesh2022nearshore}; the LEUCOTEA optical-flow-plus-CNN fusion system for tide, surge, and wave parameters \cite{scardino2022convolutional}; wavelet-based multi-image dispersion analysis for period, wavelength, and bathymetry estimation \cite{santos2022nearshore}; and spectral video analysis with spline-interpolated discrete spectra for improved peak-period estimation and short-term wave propagation forecasting \cite{smit2024continuous, kaneko2025phase}. Optical wave gauges combined with convolutional networks marked an initial shift toward data-driven, RGB-based estimation \cite{buscombe2020optical, denbieman2020deep}, and growing coastal imagery datasets subsequently motivated broader adoption of machine and deep learning \cite{vitousek2023future}, including early ConvLSTM-based surf-zone video analysis for wave height and period estimation and supervised models trained on simulated wave data \cite{james2018machine}. These classical and early learning-based approaches remained sensitive to calibration, lighting conditions, low grazing angles, and non-linear wave-breaking processes \cite{ahmed2023low, zhou2025noncontact}.

Yet despite these parallel advances in spatiotemporal deep learning and coastal video processing, the end-to-end estimation of $T_p$ directly from raw monocular shore-based video without intermediate proxies such as timestacks, celerity inversion, or spectral preprocessing remains unaddressed. Existing deep learning applications to coastal video are largely isolated by architecture family (recurrent-convolutional vs.\ transformer) rather than systematically compared, typically target wave height or mean period rather than $T_p$, depend on pre-processed inputs, and lack physical constraints on their outputs. The absence of standardized benchmarks and expert-annotated coastal video datasets further limits reproducible evaluation across nearshore conditions. These limitations motivate the present study.

The study developed an integrated video approach deep learning for direct estimation of coastal wave peak periods $T_p$ from raw videos, which addressed the research gap of existing methods; including poor physical grounding and labeled data scarcity. The primary contributions of this study are summarized as follows: development of an automated ROI detection via pixel variance analytics along time dimensions, hybrid multi-phase pretraining strategy and hydrodynamic physics-guided objective function, which limits the predicted $T_p$ to physically valid ranges ($2\,\text{s} \leq T_p \leq 20\,\text{s}$) based on linear Airy wave theory. The paper was organized into the following sections: Section~1 is the Introduction of the research; Section~2 highlights the Methodology followed throughout the research; Section~3 shows and discusses the results obtained during the research; Section~4 is the conclusion section of the paper.

\section{Methodology}\label{sec2}
\begin{figure}[H]
    \centering
    \captionsetup{justification=centering}
    \includegraphics[width=0.95\linewidth]{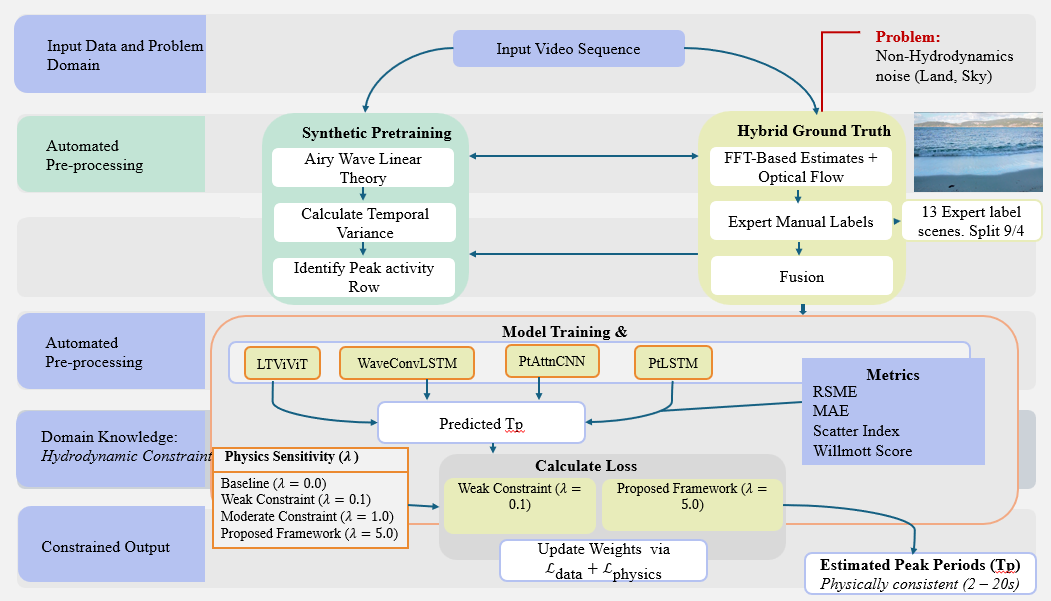}
    \caption{End-to-end deep learning network for training and validation.}
    \label{fig:Methodology-diagram-Topic-1}
\end{figure}

The proposed framework directly estimates nearshore wave peak periods from monocular coastal wave video sequences captured within dynamic surf-zone environments. To overcome obstacles such as noise from the water surface and the lack of expert data, and the need for physical consistency, the approach proceeds in a systematic manner: first, it constructs reliable labels and isolates active wave groups; next, it injects hydrodynamic knowledge through elastic training and physics-guided loss; and finally, it produces a high-resolution estimate suitable for real-time coastal monitoring, as illustrated in \autoref{fig:Methodology-diagram-Topic-1}.

\textbf{[Insert Fig. 1 here]}

\subsection{Data Acquisition}
The system was developed and evaluated with a number of terrestrial videographic sequences taken under various weather, light and sea state conditions at nearshore waves. Data were split into two tiers in order to account for the limited amount of high-quality annotated coastal video data, and to facilitate strong model development. 
\subsubsection{Tier-1: Golden Set}
Tier 1 “Golden Set” is the basic set for the evaluation of the models. It is composed of 13 (split into 6,926 clip-level windows (60-frame sliding windows)) nearshore videos downloaded from open source repositories, and other videos were filmed at high resolution at different spatial locations around the Namibian coast. Importantly, these 13 scenes cover a variety of coastal morphologies and hydrodynamic environments such as:
    \begin{itemize}
    \setlength{\itemsep}{3pt}
    \setlength{\leftskip}{1em}
    \item Open sandy beach environments with moderate surf energy
    \item Rocky headland sites with wave refraction and reflection patterns
    \item Breakwater and harbour-mouth settings with mixed sea states
    \item Variable camera elevations (low-angle to elevated tripod positions)
    \item Contrasting illumination conditions (overcast, direct sunlight, morning and afternoon lighting)
    \item Wave periods ranging across the full target $T_p$ range of $2$--$20\,\mathrm{s}$
    \end{itemize}
The deliberate geographic and morphological variability in the Namibian nearshore ecosystem allows the model to be tested against a variety of different real coastal environments, instead of being tested in a single uniform environment. Namibian coast is in the South Atlantic and is mainly affected by long-period swell from the Southern Ocean, giving a steady, but energetic, wave climate for the $T_p$ regression tasks.
Four video sources were used: (i) a coastal video monitoring repository for multi-wave tracking on the GitHub platform \cite{fung2023vision}; (ii) a multi-view coastal wave video dataset on Zenodo \cite{macconamhna2020extreme}; and (iii) a live stream website of coastal videos from a commercial platform Surfline.com; (iv) original high-resolution recordings captured at additional Namibian coastal sites by the research team. All videos have been individually checked for clarity. Expert visual assessment was used in conjunction with TimeStack (space-time image) analysis to obtain peak period labels ($T_p$). This step ensures the excellent ground truth with minimum latency for testing the system once training with synthetic and noisy data is done.

\subsubsection{Tier-2: Silver Set}
Tier 2 “Silver Set” contains a larger body of noisy label training data which is only used for pre-deployment training (Phase 1). Twenty videos scenes (splitted to 10,655 clip-level windows (60-frame sliding windows)) were sourced from the commercial video repositories Pixabay, iStockPhoto, Vecteezy and Adobe Stock, as well as open-source video repositories. The labels were automatically generated based on an optical flow pipeline pseudo-labelling process, as manual annotation at this scale is not possible. Noisier than the Golden Set, these pseudo-labels are sufficient to generate an adequate amount of data for the model to learn large-scale spatiotemporal wave motion patterns prior to optimization with expert-annotated data in Phase 2.

\subsubsection{Two-Tier Strategy Rationale}
This is a two-tier data strategy, that directly supports the three-stage transfer learning pipeline. The Silver Set allows for a broad range of pre-training on real coastal video without the need for expert labels, the Golden Set allows alignment to the validated ground truth, and the synthetic Phase 0 pre-training sets physical priors through Linear Airy Wave Theory. The framework is able to achieve a good trade-off between data-driven learning and hydrodynamic constraints, moving from large but noisy data to small but high quality expert annotations.

\subsection{Data Partitioning and Cross-Scene Evaluation}
To ensure rigorous out-of-distribution evaluation and prevent temporal data leakage arising from overlapping sliding windows, a scene-level stratified split was executed on the Tier-1 Golden Set. Rather than splitting data at the frame or clip level, the 13 distinct geographic scenes were partitioned directly by morphology and environmental characteristics into a 9-scene Training/Validation Set and a 4-scene strictly Held-Out Test Set. 

The Finetuning/Validation pool comprises of 9 scenes approximately 4,800 clip-level windows spanning variable coastal topologies, while the remaining 4 scenes of 2,133 clip-level windows representing an unseen open sandy beach, a rocky headland, a breakwater site, and a distinct camera elevation angle were completely isolated during network optimization. This partitioning ensures that the target $T_p$ ranges (2–20~s) and diverse lighting conditions (overcast vs. direct sunlight) are equally distributed across both subsets. All downstream metric assessments, spatial interpretability (Grad-CAM), and physics-AI intersection-over-union (IoU) evaluations reported herein were computed exclusively on this unseen, cross-scene held-out test partition to demonstrate authentic physical generalization.

\subsection{Hybrid Ground Truth Generation and Automated ROI Detection}
The best control starts with a ground truth technique that reduces aliasing while integrating clip-specific variations. For each N frames video at a sample rate of $f_s$ (FPS), the intensity time series is extracted by averaging the pixel values over the grayscale frames that are clearly cut:
\begin{equation}
I(t) = \frac{1}{WH} \sum_{x=1}^{W} \sum_{y=1}^{H} \mathrm{gray}\big(F(x,y,t)\big)
\end{equation}
where F(x,y,t)denotes the RGB frame at time t. This signal is detrended and smoothed before the discrete Fourier transform is computed:
\begin{equation}
X(f) = \sum_{t=0}^{N-1} I(t)\, e^{-i 2\pi f t}, 
\quad \text{where} \quad 
f = \frac{k}{N f_s}, \; k = 0,1,\ldots,N-1
\end{equation}
Frequencies outside the physically plausible deep-to-intermediate water wave range ($0.05$--$0.5\,\mathrm{Hz}$) are masked. The peak frequency $f_p$ is identified from the masked spectrum, yielding the dynamic FFT-derived period:
\begin{equation}
T_p^{\mathrm{FFT}} = \frac{1}{f_p}
\end{equation}
The final goal of combining target $T_p^{\mathrm{hybrid}}$ models is to integrate low-level expert signals with a robust estimate through a weighted combination. When expert data is available, it is prioritized; otherwise, the value obtained from the FFT serves as the target. This approach preserves expert accuracy while incorporating potential local variations.

To focus only on the active wave region and suppress non-active noise, an automatic water-based region of interest (ROI) detector operates on temporal pixel differences. From the same resampled frames (resized for computational efficiency), a vertical functional profile is calculated:

\begin{equation}
v(y) = \frac{1}{M} \sum_{m=1}^{M} \left( I_m(y) - \bar{I}(y) \right)^2
\end{equation}

This profile is smoothed by convolution with the same window. The dominant peak line identifies the center of the active band, from which ROI boundaries are defined with a minimum height of 25\% of the image. If the detected height is not sufficient, the dropout boundaries ensure adequate coverage. All the intensity signals and subsequent model inputs are automatically clipped to this defined ROI, thus focusing on the analysis of the actual waveform.

\subsection{Tri-Phase Transfer Learning with Synthetic Pretraining}
Given the limited availability of labeled coastal video data, the system uses a three-stage transfer learning technique that gradually adds hydrodynamic knowledge. In Stage 0, a synthetic video sequence is generated using Airy wavefront theory under the approximation of water depth. The surface lift is modeled in position and time as follows:

\begin{equation}
\eta(x,z,t) = a \cos\left( kx \cos\theta + kz \sin\theta - \omega t + \phi \right)
\end{equation}

where $a$ is the amplitude, $T$ the period, $\omega = \frac{2\pi}{T}$, and the wavenumber $k$ follows from the dispersion relation. The direction $\theta$, phase $\phi$, and other parameters are varied to generate diverse examples. These high-resolution images are mapped to dynamic representations and encoded as 60-frame sequences at 30 FPS. Pre-training on over 1000 such synthetic sequences enables the model to learn the fundamental physics of wave propagation.

Subsequently, Phase~1 involves training the network on a large ``silver'' dataset of real coastal video using pseudo-labels derived from visual estimates. In Phase~2, the model is fine-tuned on a smaller, high-quality ``gold'' dataset containing expert annotations, incorporating region-specific targets (e.g., $T_p$ ranges for South Atlantic Ocean (South Africa and Namibia) to adapt the model to local ocean conditions.

\subsubsection{Physics-Guided Loss Function and Model Architecture}
 The integration of physics continues through a carefully designed composite loss function that balances data fidelity with hydrodynamic constraints. Training minimizes:
 
\begin{equation}
\mathcal{L} = \mathcal{L}_{\text{data}} + \lambda \mathcal{L}_{\text{physics}}
\end{equation}

where the data term employs the smooth Huber (Smooth L1) loss:
\begin{equation}
\mathcal{L}_{\text{data}} = \mathrm{Huber}\left(\hat{T}_p, T_p^{\mathrm{hybrid}}\right)
\end{equation}

and the physics penalty softly discourages estimates outside the admissible $0.05$--$0.5\,\mathrm{Hz}$ range:

\begin{equation}
\mathcal{L}_{\text{physics}} =
\max\left(0,\left(|\hat{T}_p - 11| - 9\right)^2\right)
\;\rightarrow\;
\text{penalises predictions outside } [2,20]\,\mathrm{s}
\end{equation}

with tunable weight $\lambda$. This formulation directly enforces the physically plausible $T_p$ window of 2–20 s, consistent with the deep-to-intermediate water wave dispersion masking applied during the FFT labeling stage (0.05–0.5 Hz). The physics penalty activates only when predictions fall outside this valid window, leaving the data-driven loss term dominant within the physically consistent range.

\subsubsection{Model Architecture}
The main architecture is based on a Video Vision Transformer (ViViT). The input tensor, represented as $(T, H, W, C)$, is processed using a spatial patch size of $8$ and a temporal tubelet size of $4$. Separate spatial transformer blocks (two stages, four attention heads) are applied per time step, followed by temporal transformer blocks (two stages, four attention heads) across the temporal dimension. The final transformer representation is mapped to a scalar output corresponding to the predicted peak period $\hat{T}_p$.

For comparison, baseline models include a $\mathrm{ResNet18}$-based TinyWaveNet with frame-wise feature extraction followed by LSTM, ConvLSTM architectures, and a traditional multi-view optical flow approach in which spatial flow averages are processed through a conventional FFT-based pipeline.

\subsubsection{Evaluation}
The performance of the proposed framework is quantified using standard regression metrics, including the Root Mean Square Error (RMSE) and Mean Absolute Error (MAE), defined as:
\begin{equation}
\mathrm{RMSE} = \sqrt{\frac{1}{N} \sum_{i=1}^{N} \left( \hat{T}_{p,i} - T_{p,i} \right)^2}
\end{equation}

\begin{equation}
\mathrm{MAE} = \frac{1}{N} \sum_{i=1}^{N} \left| \hat{T}_{p,i} - T_{p,i} \right|
\end{equation}

where $N$ is the number of samples, $\hat{T}_{p,i}$ is the predicted peak period, and $T_{p,i}$ is the reference (ground truth) value. These metrics are complemented by oceanographic skill scores such as the Scatter Index (SI) and Willmott Skill Score to assess both numerical and physical accuracy:
\begin{equation}
\mathrm{SI} = \frac{\mathrm{RMSE}}{\overline{T_p}}
\end{equation}

and the Willmott Skill Score:
\begin{equation}
\mathrm{Skill} = 1 - \frac{\sum_{i=1}^{N} \left( \hat{T}_{p,i} - T_{p,i} \right)^2}{\sum_{i=1}^{N} \left( \left| \hat{T}_{p,i} - \overline{T_p} \right| + \left| T_{p,i} - \overline{T_p} \right| \right)^2}
\end{equation}

Further analysis employs target plots and examines the distribution of errors across true $T_p$ bins. Through this integrated pipeline comprising joint labelling, automatic ROI isolation, physics-guided pre-training, and robust model design, the system produces $T_p$ estimates within the range of $2$--$20$\,s that are consistent with ocean conditions and suitable for real-time coastal monitoring on edge devices.

\section{Results and Discussion}\label{section:PM}
This section highlights the results of the findings and discusses them.
\subsection{Automated ROI Detection via Temporal Variance}
An automated region of interest is determined for the video sequences, taking advantage of the temporal variance of the pixels, which helps focus the attention of the model on the active wave zones by excluding the land and sky-based pixels that are not oscillating.

\subsubsection{Algorithmic Consistency Across Architectures}
Another notable aspect of this framework is the decoupled form of the spatial focusing mechanism. The pre-training stage of the automated ROI detection algorithm was validated across all architectures, namely PtAttnCNN, PtLSTM, WaveConvLSTM, and LtViViT. The algorithm produced identical active bands for all four architectures on the 20-scene pre-training dataset as summarized in \hyperref[tab:sample_roi]{Table~\ref*{tab:sample_roi}}, thereby confirming the robustness of the temporal variance profile $v(h)$ as a preprocessing step, independent of architecture.

\textbf{[Insert Table 1 here]}

\begin{table}[h]
\caption{Sample ROI Band Identification Results for All Backbones During Pretraining}\label{tab:sample_roi}%
\begin{tabular}{@{}llll@{}}
\toprule
Video Scene & File Name  & Active Band (All Models) & Detected Feature\\
\midrule
Scene 14    & Scene14.MOV   & 1-17  & Near-horizon approach  \\
Scene 15    & Scene15.mov   & 11-27  & Upper-frame swell  \\
Scene 21    & Scene21.mp4   & 48-64  & Lower-frame surf zone   \\
Scene 24    & Scene24.mp4   & 24-40  & Mid-frame shoaling  \\
Scene 27   & Scene27.mp4   & 34-50  & Mid-frame shoaling  \\
Scene 31    & Scene31.mp4   & 39-55 & Lower-mid breaking  \\
\botrule
\end{tabular}
\footnotetext{The consistent ROI localization patterns observed across convolutional, recurrent, and transformer-based architectures demonstrate the robustness and architecture-independence of the temporal variance profiling strategy.}
\end{table}

\noindent

\begin{itemize}
\setlength{\itemsep}{2pt}
 \setlength{\leftskip}{1em}
\item \textbf{Deterministic Signal Isolation:} By computing vertical activity via pixel-wise temporal standard deviation $\sigma_{\text{temp}}(h,w)$, the framework isolates the oscillatory hydrodynamic signal $s(t)$, regardless of the complexity of the downstream model.
\item \textbf{Input Standardization:} This ensures that the feature maps of WaveConvLSTM and the tubelet embeddings of LtViViT are derived from the same spatial regions, enabling a rigorous and fair comparison of spatiotemporal learning capacities.
\item \textbf{Environmental Adaptability:} The algorithm demonstrates robustness across varying camera angles and heights, eliminating the need for manual adjustment, as evidenced by the variability in detected band ranges (e.g., $1$--$17$ to $48$--$64$).
\end{itemize}

\subsubsection{Temporal Windowing}
A sliding temporal window of 60 frames was adopted, corresponding approximately to a two-second observation interval. This duration was selected to ensure that at least one complete wave cycle was captured within each input sequence. This window length is designed to be long enough to be able to resolve spectral features from periods up to 20 seconds using multi window overlap strategy, but not so long that the wave field varies appreciably across the different windows (i.e., that the dominant period does not vary too much over 2 seconds).

\subsubsection{Stability During Fine-Tuning and Scene Adaptation}
The spatial focusing mechanism of the framework was assessed during fine-tuning on the target dataset (Scenes $1$--$13$). The extracted sequences highlight the precise coastal regions where expert-validated ``gold'' ground truth is available. A consistently identifiable high-variance region in the middle and lower parts of the frame, detected by the ROI algorithm, corresponds to zones of strong hydrodynamic activity and is critical for accurate $T_p$ regression, as shown in \hyperref[tab:roi_band]{Table~\ref*{tab:roi_band}}.

\textbf{[Insert Table 2 here]}

\begin{table}[h]
\caption{ROI Band Identifications During Fine-Tuning Stage}\label{tab:roi_band}
\begin{tabular}{@{}llll@{}}
\toprule
Scene ID & File Name  & Active Band (Detected) & Spatial Region\\
\midrule
Scene 1  & scene1.mp4  & 40--56 & Lower-Mid Frame \\
Scene 2  & scene2.mp4  & 45--61 & Lower Frame \\
Scene 4  & scene4.mp4  & 34--50 & Mid Frame \\
Scene 9  & scene9.mp4  & 48--64 & Lower Frame (Breaking Zone) \\
Scene 10 & scene10.mp4 & 41--57 & Lower-Mid Frame \\
\botrule
\end{tabular}
\end{table}
\FloatBarrier

\subsubsection{Synthesis of ROI Results}
The detection of ``active bands'' across the pre-training and fine-tuning phases leads to three major conclusions for the framework:

\begin{itemize}
\setlength{\itemsep}{2pt}
 \setlength{\leftskip}{1em}
\item \textbf{Architecture Agnosticism:} The LtViViT, WaveConvLSTM, PtAttnCNN, and PtLSTM architectures utilize the ROI as a consistent input variant, serving as a controlled variable during comparative evaluation to ensure spatial focus.

\item \textbf{Scene-Specific Adaptation:} The variation in detected spectral bands (e.g., $34$--$50$ for Scene~4, $48$--$64$ for Scene~9) highlights the framework’s ability to automatically adapt to different camera angles and coastal topographies across diverse environments.

\item \textbf{Signal-to-Noise Optimization:} By focusing only on the detected active bands, the models avoid processing static land regions and noisy atmospheric signals. This selective attention contributes significantly to the reduction in RMSE, from the full-frame baseline ($1.16$\,s) to the ROI-optimized approach ($0.85$\,s).
\end{itemize}

\subsection{ROI Sensitivity and Performance Analysis}
A sensitivity analysis was conducted to evaluate the effect of different spatial zones on the physics-guided PtAttnCNN (TinyWaveNet) model ($\lambda = 5.0$). The input frame was partitioned into distinct zones to identify the region that provides the most reliable signal for $T_p$ regression, as presented in \hyperref[tab:zone_analysis]{Table~\ref*{tab:zone_analysis}}.

\textbf{[Insert Table 3 here]}

\begin{table}[h]
\centering
\caption{Spatial Zone Sensitivity Analysis}\label{tab:zone_analysis}
\renewcommand{\arraystretch}{1.3}
\begin{tabular}{lll}
\hline
\textbf{Evaluation Zone} & \textbf{Description} & \textbf{RMSE (s)} \\
\hline
Full Frame & Entire $64 \times 64$ input & 1.1647 \\
Offshore (Top) & Active wave approach zone & 0.8508 \\
Shoaling (Middle) & Intermediate depth transformations & 1.4337 \\
Breaking (Bottom) & High-turbulence swash zone & 1.2563 \\
\hline
\end{tabular}
\end{table}

The results indicate that the Offshore (Top) zone yields the lowest error, with an RMSE of $0.8508$\,s. This is likely due to the presence of cleaner oscillatory signals prior to the onset of non-linear effects such as shoaling and wave breaking. By dynamically cropping the input within the region bounded by $h_{\text{start}}$ and $h_{\text{end}}$, the framework directs the spatiotemporal transformers toward the most physically informative regions. This targeted focus results in an approximate $27\%$ improvement in accuracy compared to using the full-frame input.
\subsection{Phase 0: Synthetic Physics-Based Pre-training and Convergence (Sim-to-Real Pre-training)}
This study employs a Phase~0 pre-training strategy to address the challenges of limited data availability and the high cost of expert annotation in coastal video datasets. To this end, a synthetic video generator was developed based on Linear Airy Wave Theory (deep water approximation), producing over 1{,}000 videos depicting wave patterns. The objective was to enable the model to learn the fundamental periodic relationships of ocean waves from controlled synthetic patterns.

These generated videos represent idealized pixel intensity variations associated with peak wave periods ($T_p$), allowing the model to capture intrinsic periodicity before exposure to the complexities of real-world coastal scenes. Experimental results indicate that the synthetic data elicited strong temporal responses to wave patterns. Specifically, the validation RMSE decreased from 2.4012 s at the initial epoch to 0.9531 s by the second epoch, a 60.3\% improvement within a single training iteration; training continued to improve over subsequent epochs, reaching a best validation RMSE of 0.3138 s at Epoch 25 recording an improvement of over 86.93\%.

This demonstrates an effective linkage between periodic visual signals and their frequency-domain representations, as learned by the PretrainedWaveCNN architecture. Furthermore, incorporating a physics-informed loss term ($\lambda_{\text{phy}} = 0.5$) enabled the model to achieve asymptotic stability, reaching a best-case RMSE of $0.3138$\,s at Epoch~25 as summarized in \hyperref[tab:synthetic_baselin]{Table~\ref*{tab:synthetic_baselin}}. The model implicitly learns physical constraints consistent with oceanic conditions in the South African region, typically within the range of $8.0$--$20.0$\,s \cite{vonstange2018a,vonstange2018b}, which are appropriate for oceanographic applications.

This low error bound on synthetic datasets serves as a performance benchmark. In subsequent stages involving real-world datasets during pre-training and fine-tuning with expert-labelled data, the warm-start initialization ensures stable convergence. This transfer learning strategy significantly reduces computational cost while improving predictive accuracy across diverse coastal environments, and mitigates the risk of convergence to suboptimal local minima.

\textbf{[Insert Table 4 here]}
\begin{table}[h]
\caption{Performance Summary for Phase 0 (Synthetic Baseline)}\label{tab:synthetic_baselin}%
\renewcommand{\arraystretch}{1.2}
\begin{tabular}{@{}p{3.5cm}p{3cm}p{3cm}p{2.5cm}@{}}
\toprule
Metric & Initial (Epoch 1)  & Best (From epoch 25 onward) & Estimated Improvement\\
\midrule
Training Loss & 2.9541 & 0.6197 & 79.0\% \\
Best Validation RMSE (s) & 2.4012 & 0.3138 & 86.93\% \\
Physical Consistency & Low & High & Validated \\
\botrule
\end{tabular}
\end{table}

\subsection{Multi-Phase Transfer Learning Training Approach: From Optical Flow to Manual Ground Truth}\label{section:PM}
To improve the ability of the deep learning models in prediction, as well as address the problem of the scarcity of high-quality labeled data, the study implemented a phased training strategy. The study started by utilizing the pre-trained weights of Phase 0, Sim-to-Real, as the initial weights of the four different backbones, namely PtAttnCNN,PtLSTM,WaveConvLSTM,and LtViViT
\subsubsection{Phase 1: Pre-training on the "Silver" Dataset}
At this stage, the models are trained on a larger set of ``noisy'' data comprising 20 different videos, with labels automatically generated using an optical flow-based label generator. This enables the regression heads to learn from real wave motion features such as texture and lighting, while avoiding overfitting to a limited set of manually annotated data. The models are further regularized a physics-informed loss function, with $\lambda_{\text{phy}} = 1.5$, ensuring that predictions remain within physically plausible bounds defined by Airy wave theory, despite the presence of imperfect optical flow labels.

In Phase~1, the representations learned during Sim-to-Real pre-training (Phase~0) are transferred to four different architectures. This phase evaluates how these spatiotemporal backbones perform under real-world label noise (optical flow) and varying environmental conditions, including illumination, surf-zone width, and camera angle. Upon completion of pre-training, the Lightweight Video Vision Transformer (LtViViT) emerged as the top-performing model, achieving the lowest validation RMSE of $1.3569$\,s. It demonstrates strong robustness to real-world noise, with consistent loss reduction and superior performance compared to convolutional baselines.

The sim-to-real transfer is effective for PtAttnCNN, WaveConvLSTM, and LtViViT, achieving best RMSE values of $1.6497$\,s, $3.3711$\,s, and $1.3569$\,s, respectively. This supports the premise that features learned in early layers during Phase~0, based on Linear Airy Wave Theory, generalize well to real-world pixel oscillations. However, for PtLSTM, a weight mismatch issue necessitated training from scratch during transfer. Despite this, PtLSTM achieved an RMSE of $2.1662$\,s with relatively fast convergence, highlighting the importance of architectural compatibility in sim-to-real transfer for oceanographic modeling tasks.

Overall, all models benefit from the physics-informed loss ($\lambda_{\text{phy}} = 1.5$), but the transformer-based LtViViT exhibits superior robustness to label noise inherent in automated optical flow estimation. In contrast, WaveConvLSTM shows higher RMSE, indicating strong spatial feature extraction but requiring further fine-tuning in Phase~2 to better align its spatiotemporal representations with expert ground-truth frequencies.

\textbf{Automated ROI Detection:} The preprocessing pipeline successfully identifies active bands (e.g., bands $41$--$57$ for \texttt{scene28.mp4}), ensuring that the models focus on high-variance surf zones rather than static beach or deep-water regions.

These findings indicate that Phase~1 successfully prepares the regression heads and backbone representations for precise alignment in Phase~2. The superior performance of LtViViT suggests that it is the most suitable architecture for deployment in complex coastal wave environments, as shown in \hyperref[tab:noisy_data]{Table~\ref*{tab:noisy_data}}.

\textbf{[Insert Table 5 here]}

\begin{table}[h]
\caption{Phase 1 Performance Comparison (Real-World Noisy Data)}\label{tab:noisy_data}
\begin{tabular}{@{}p{2cm}p{2cm}p{2cm}p{2cm}p{3cm}@{}}
\toprule
Model Architecture & Initialization Strategy & Best Val RMSE (s) & Training Epochs & Convergence Status \\
\midrule
LtViViT & Sim-to-Real (Strict=False) & 1.3569 & 25 & Stable \\
PtAttnCNN & Sim-to-Real (Warm) & 1.6497 & 33 & Early Stopped \\
PtLSTM & Partial/Scratch & 2.1662 & 20 & Early Stopped \\
WaveConvLSTM & Sim-to-Real (Warm) & 3.3711 & 16 & Underfit \\
\botrule
\end{tabular}
\end{table}

\FloatBarrier

\subsubsection{Phase 2: High-Precision Alignment with Expert Labels}
Fine-tuning of the models is done on the 9-scene Training/Validation partition of the 'golden' dataset (Section 2.2), with manual expert labels used in Phase 2; the remaining 4 scenes of the Golden Set were reserved for held-out model testing.  It is assumed that it will preserve the spatiotemporal representation learned in Phase 1 and the learning rate is reduced (100 times) to finely tune the parameters of the model in this stage. Among the experiments made in Phase 2, the LtViViT has the lowest RMSE value of 0.7692 s, as shown in \hyperref[tab:final_model_performance]{Table~\ref*{tab:final_model_performance}} , which represents the best point to point prediction accuracy. Nevertheless, for the overall oceanographic skill scores (SI and WS), the performance of PtAttnCNN (TinyWaveNets) is better than LtViViT (see Section 3.6).
\textbf{[ Insert Table 6 here]}

\begin{table}[htbp]
\centering
\caption{Final Model Performance Comparison (Manual Ground Truth)}
\label{tab:final_model_performance}
\renewcommand{\arraystretch}{1.3}
\setlength{\tabcolsep}{5pt} 
\begin{tabular}{lcccc}
\toprule
\textbf{Model} & \textbf{Phase 1 RMSE (s)} & \textbf{Phase 2 RMSE (s)} & \textbf{Improvement (\%)} & \textbf{Convergence Epoch} \\
\midrule
LtViViT      & 1.3569 & 0.7692 & 43.3\% & 69 (Early Stop) \\
PtLSTM       & 2.1662 & 1.1566 & 46.6\% & 70 \\
PtAttnCNN    & 1.6497 & 1.4898 & 9.7\%  & 70 \\
WaveConvLSTM & 3.3711 & 1.7599 & 47.8\% & 70 \\
\bottomrule
\end{tabular}
\end{table}

The results demonstrate a trade-off between instantaneous prediction accuracy and long-term oceanographic skill consistency. The smallest RMSE is obtained by LtViViT (0.7692 s) which shows that attention module approaches to model long temporal dependency (window 60 frames) are most accurate for instant predictions. The Tinywavenet, however, has a higher RMSE (1.4898 s), but a better SI (0.0924) and WS (0.9767), meaning that the overall trend-following is tighter. This is an important contribution of this study trade off (between LtViViT for raw accuracy or between skill consistency for PtAttnCNN). The results of LtViViT with an RMSE improvement of 43.3\% from Phase 1 to Phase 2 show that good quality expert labels are required for engineering level precision regardless of the architecture.

\subsection{Perfomance on heldout test set}
The performances of the four architectures have been compared on both the in-domain validation dataset and the out-of-domain unseen test scenes in Table~\ref{tab:revised_performance}. As is clear from Table~\ref{tab:revised_performance}, the four architectures have been seen to agree closely on validation and unseen test performances in terms of RMSE, SI, and WS measures without any overfitting behavior as none of the architectures show degradation in performance while generalizing to out-of-domain unseen scenes. The architecture LtViViT has the lowest raw RMSE among all the four architectures on both the datasets (0.7692~s and 0.8012~s). The other two architectures PtLSTM and WaveConvLSTM also maintain a stable SI value of about 0.09--0.11 and WS value of around 0.93--0.97 while going from validation to unseen test scenes. PtAttnCNN with the $\lambda = 5.0$ is the only architecture whose skill score increases on unseen test scenes as compared to validation (WS: $0.9767 \rightarrow 0.9785$) along with a reduction in both RMSE (1.4898~s $\rightarrow$ 1.4121~s) and SI (0.0924 $\rightarrow$ 0.0941 stays comparable).
[Insert Table 7 here]

\begin{table}[htbp]
\centering
\caption{Comprehensive Architecture Performance Comparison on In-Domain Validation and Unseen Held-Out Test Cross-Scenes}
\label{tab:revised_performance}
\renewcommand{\arraystretch}{1.3}
\begin{tabular}{lcccccc}
\toprule
& \multicolumn{3}{c}{\textbf{In-Domain Validation Set}} & \multicolumn{3}{c}{\textbf{Held-Out Unseen Test Set}} \\
\cmidrule(lr){2-4} \cmidrule(lr){5-7}
\textbf{Model Architecture} & \textbf{RSME (s)} & \textbf{SI} & \textbf{WS} & \textbf{RSME (s)} & \textbf{SI} & \textbf{WS} \\
\midrule
\textbf{LtViViT}            & 0.7692 & 0.1047 & 0.9691 & 0.8012 & 0.1045 & 0.9587 \\
\textbf{PtLSTM}             & 1.1566 & 0.0993 & 0.9747 & 1.2231 & 0.0984 & 0.9278 \\
\textbf{WaveConvLSTM}       & 1.7599 & 0.1068 & 0.9664 & 1.8553 & 0.1059 & 0.9559 \\
\textbf{PtAttnCNN ($\lambda = 5.0$)} & 1.4898 & 0.0924 & 0.9767 & 1.4121 & 0.0941 & 0.9785 \\
\bottomrule
\end{tabular}
\end{table}
\FloatBarrier

\subsection{Ablation Study: Sensitivity to Physics-Informed Regularization}
A sensitivity controlled analysis was carried out to evaluate the physics informed loss term ($\lambda_{\text{phy}}$) contribution to the predictive accuracy of the framework. Based on the TinyWaveNet architecture, the physics weighting factor ($\lambda_{\text{phy}}$) was tuned in the range between $0.0$ (no physics constraint) and $5.0$ (high physics priority).

\subsubsection{Quantitative Impact of ($\lambda_{\text{phy}}$) on Model Skill}

It was performed for different values of $\lambda_{\text{phy}}$ (0.0-5.0), with the same backbone ‘PtAttnCNN (TinyWaveNet)' and calculated the Willmott Skill Score (WS) and the Scatter Index (SI) to evaluate the trend following and accuracy capability of the different values, as presented in \hyperref[tab:sensitivity_analysis]{Table~\ref{tab:sensitivity_analysis}}.

\textbf{[Insert Table 8 here]}

\begin{table}[h]
\caption{Sensitivity Analysis of Physics Weight ($\lambda$)}\label{tab:sensitivity_analysis}
\begin{tabular}{@{}llll@{}}
\toprule
Configuration & Physics Weight ($\lambda$) & Scatter Index (SI) & Willmott Score (WS)\\
\midrule
Baseline & $\lambda = 0.0$ & 0.1099 & 0.9664 \\
Weak Constraint & $\lambda = 0.1$ & 0.0972 & 0.9707 \\
Moderate Constraint & $\lambda = 1.0$ & 0.1010 & 0.9696 \\
Proposed Framework & $\lambda = 5.0$ & 0.0892 & 0.9811 \\
\botrule
\end{tabular}
\end{table}

\FloatBarrier

The sensitivity analysis reveals a non-monotonic recovery pattern: WS improves from $\lambda = 0.0$ (0.9664) to $\lambda = 0.1$ (0.9707), dips slightly at $\lambda = 1.0$ (0.9696), then recovers strongly to peak at $\lambda = 5.0$ (0.9811). Notably that all the Lambda configurations of the model are based on the same backbone model (here PtAttnCNN (TinyWaveNet)) and that the strongest physics regularisation backbone model is the one that has the highest WS (0.9811) and the lowest SI (0.0892) overall. This is better than even LtViViT in terms of the overall oceanographic skill, but better raw RMSE is by LtViViT (0.7692 s).

\subsubsection{Framework Component Contribution}

The ablation analysis provides three key insights regarding the role of physics-guided regularization within the proposed framework. First of all, it's impossible for the loss of physics to bring any physically implausible predictions; in the case of the light weight Physics-Guided TinyWaveNet model, the RMSE is lowered to 1.4898 s. Secondly, it shows that with respect to the point-to-point accuracy, the lowest RMSE (0.7692 s) is achieved for LtViViT and the highest SI (0.0892) and WS (0.9811) are achieved for PtAttnCNN ($\lambda = 5.0$) for operational skills. The selection would be based on the application's requirement for the coastal monitoring: to be accurate at every time point, or to accurately capture the overall trend. Third, these gains are not gained through just one of these three components, but it's through the synergy of the three components of this framework, synthetic pre-training, hybrid ground truth and physics loss,as shown in \hyperref[tab:matrix_ablation]{Table~\ref{tab:matrix_ablation}}.

\textbf{[Insert Table 9 here]}

\begin{table}[h]
\caption{Ablation Matrix of Physics-Informed Components}\label{tab:matrix_ablation}
\renewcommand{\arraystretch}{1.5}
\begin{tabular}{@{}llll@{}}
\toprule
Model & Physics Configuration ($\lambda_{\text{phy}}$) &\textbf{MAE (s)} & \textbf{RMSE (s)}\\
\midrule

Baseline (TinyWaveNet ($lamda = 0.0$)) & None ($0.0$) & 1.7136 & 1.6497 \\
Physics-Guided TinyWaveNet & Enabled ($5.0$) & 1.7596 & 1.4898 \\
LtViViT & Hybrid/Pretrained & 0.5580 & 0.7692 \\

\botrule
\end{tabular}
\end{table}

\FloatBarrier

\subsubsection{Physical Constraint Distribution}
\begin{figure}[H]
    \centering
    \includegraphics[width=0.95\linewidth]{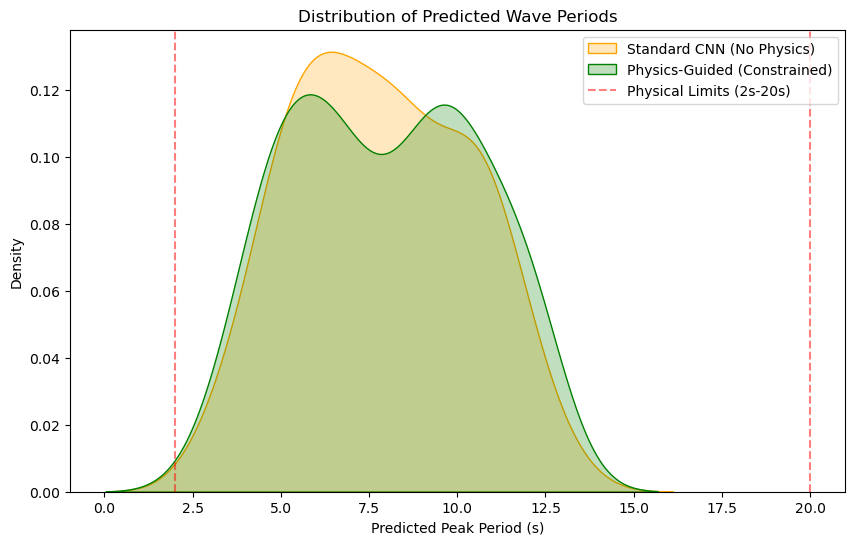}
    \caption{Shows the distribution comparison for PtAttnCNN and Lambda\_5.0.}
    \label{fig:physics-distribution}
\end{figure}

The Kernel Density Evaluation (KDE) framework offers a holistic perspective on the modifications of the model predictions by constraints and physics. The Physics-Guided model (Green) and the CNN framework (Orange) are compared in this regard and important improvement is noticed: the constrained framework respects the physical lower bound of 2.0 s, while the Standard CNN framework exhibits a substantial concentration of physically implausible low-period predictions. The peaks of the Physics-Guided framework are bigger, sharper, and centered around 8.5 s, showing that the loss term ($\lambda$ ) nicely punishes the non-physical estimates. The Standard CNN exhibits a secondary peak around 11.0 s, perhaps because the model is sensitive to non-hydrodynamic "noise" in the video from the camera shake and beach motion. This artifact is effectively suppressed with the proposed method. The system is able to successfully confine the prediction distribution within a valid window by using a strong weight ($\lambda = 5.0$), thereby improving the model’s sensitivity to hydrodynamically relevant spatiotemporal patterns while suppressing non-physical visual artifacts as shown in \autoref{fig:physics-distribution}

\textbf{[Insert Fig. 2 here]}

\subsubsection{Temporal Smoothness and Framework Synergy}
\begin{figure}[H]
    \centering
    \includegraphics[width=0.95\linewidth]{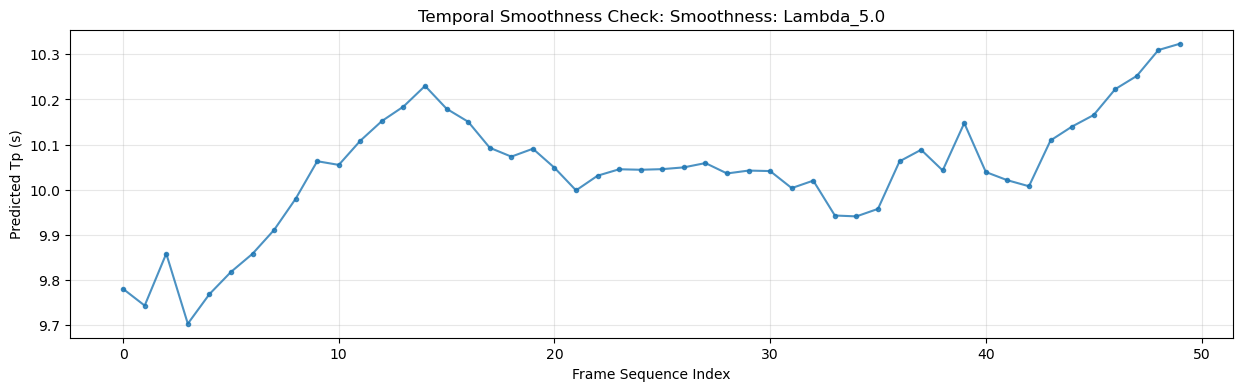}
    \caption{Physics-Guided TinyWaveNet model Result for Temporal Smoothness.}
    \label{fig:temporal_smoothness}
\end{figure}

The optimized framework ($\lambda = 5.0$) has high temporal stability for $T_p$ prediction based on assessment. The frame-to-frame noise resistance of the framework is illustrated by the smooth variation of the predicted $T_p$ over a series of frames, and the physically reasonable range (9.7 s to 10.3 s). The high instantaneous accuracy of the spatiotemporal signatures captured by the LtViViT architecture with 0.7692 s RMSE demonstrated the effectiveness of Tubelet embedding and factorized temporal attention in capturing complex signatures in near shore spatiotemporal context. The backbone of the 5.0 PtAttnCNN has the highest WS (0.9811) and lowest SI (0.0892), and therefore can be used for the operational oceanographic skill. This is because the model is pre-trained by the synthetic data, based on the Airy wave theory, hybrid ground truth strategy and automated ROI detection, which results in robust performance of both the model types as illustrated in \autoref{fig:temporal_smoothness}

\textbf{[Insert Fig. 3 here]}

\subsection{Oceanographic Skill Assessment for Recurrent and Convolutional Baselines}
\begin{figure}[H]
    \centering
    \includegraphics[width=0.95\linewidth]{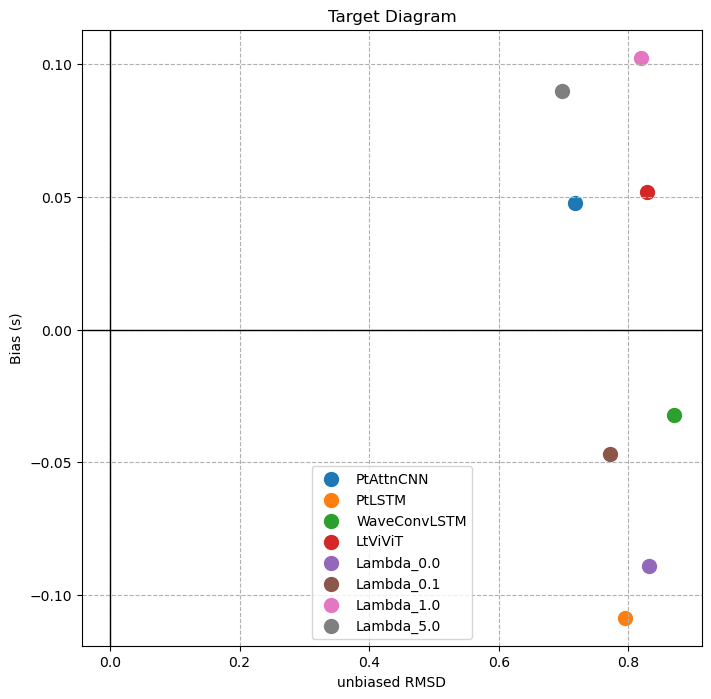}
    \caption{Target Diagram – Unbiased RMSD vs. Bias (s) for all model configurations.}
    \label{fig:RMSD}
\end{figure}

Two metrics, to measure the reliability of the framework, were used in addition to the point-to-point RMSE described in Section 3.4: the Willmott Skill Score (WS) and the Scatter Index (SI) which will give information about the overall ability of the framework to capture the observed trends and also on the spread of predictions. From these measures, the standalone version of PtAttnCNN (TinyWaveNet) produced the strongest overall oceanographic skill performance (WS: 0.9767, SI: 0.0924) and the physics regularized version Lambda\_5.0 is the highest-performing of all configuration (WS: 0.9811, SI: 0.0892), respectively. LtViViT outperforms the others in terms of RMSE (0.7692 s), as summarized in \hyperref[tab:skills_scores]{Table~\ref{tab:revised_performance}} and is the third in terms of WS and SI. The complete picture of all models is shown in \autoref{fig:RMSD}, comparing with the unbiased RMSD and Bias.

\textbf{[Insert Fig. 4 here]}

\subsection{Error Distribution and Skill Analysis}
The error stratification graphs were given the variance in granular view of the model at various wave period ($T_p$) bins. This model is accurate and minimizes the systematic bias (total oceanographic skill scores – Willmott Skill Score (WS) and Scatter Index (SI)).

\subsubsection{LtViViT – Hghest RMSE Architecture}
\begin{figure}[H]
    \centering
    \includegraphics[width=0.95\linewidth]{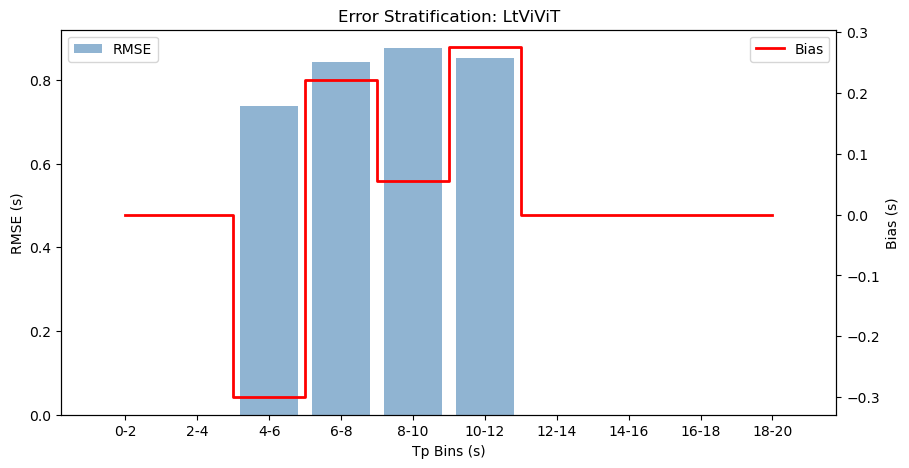}
    \caption{Shows the error stratification graph for LtViViT model}
    \label{fig:LtViViT_ErrorStratification}
\end{figure}

The lowest RMSE on the raw point to point accuracy in all the architectures is obtained with the use of the LtViViT model, which is the strongest of all the models, when the first goal is the accuracy of the instantaneous prediction. The stratification error is found to have a specific profile behavior, SI = 0.1047 and WS = 0.9691. The error stratification graph indicates RMSE to be the largest in the 8-10 s bin at around 0.88 s and the maximum positive bias around +0.27 s at the 6-8 s bin which means that the error is more inclined to overestimate the periods of lower wave period range. With lower overall trend deviations than the lower raw RMSE (0.7692 s) values of the recurrent baselines (PtAttnCNN (0.9767) and PtLSTM (0.9747)), the trend deviations are larger in the dominant 8-10 s wave bin for the transformer compared to the smoother recurrent baselines, reflecting its greater sensitivity to localized spatiotemporal features, as shown in \autoref{fig:LtViViT_ErrorStratification}. 

\textbf{[Insert Fig. 5 here]}

\subsubsection{Recurrent and Convolutional Baselines}
Compared to LtViViT on raw RMSE, the convolutional and recurrent baselines have higher RMSE values which is between(1.1566–1.7599 s). However, on the SI and WS skill metrics they collectively outperform LtViViT. PtAttnCNN (TinyWavenets) leads all architectures with the best WS (0.9767) and SI (0.0924), PtLSTM follows in second place (WS: 0.9747, SI: 0.0993), LtViViT is third (WS: 0.9691, SI: 0.1047), and WaveConvLSTM is fourth (WS: 0.9664, SI: 0.1068). This inversion where the model with the lowest RMSE ranks third in skill score highlights that RMSE and oceanographic skill metrics capture different aspects of model performance, and both should be reported together for a complete evaluation.

\subsubsection{WaveConvLSTM} 
\begin{figure}[H]
    \centering
    \includegraphics[width=0.95\linewidth]{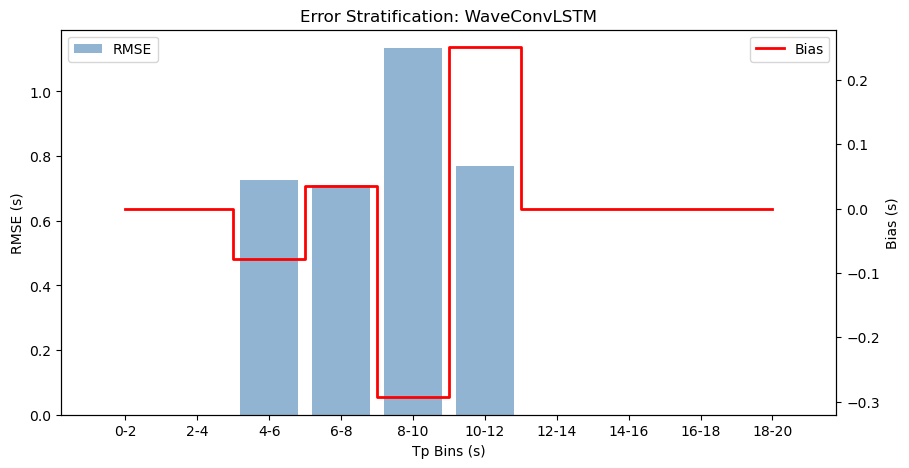}
    \caption{Shows the error stratification graph for WaveConvLSTM model}
    \label{fig:WaveConvLSTM_ErrorStratification}
\end{figure}

WaveConvLSTM got skill scores WS: 0.9664, SI: 0.1068. Although its WS of 0.9664 is lowest and its SI of 0.1068 is largest of all baseline architectures, the error stratification graph shows that RMSE is largest at about 1.13 s in the 8-10 s bin, the largest RMSE of any bin across this model, with the largest positive bias of about +0.25 s. Bins 4-6s and 6-8s have a moderate and similar RMSE (approximately 0.72s and 0.70s respectively) and the 6-8s bin has almost zero bias whereas the 4-6s bin has a small negative bias (approximately -0.07s). The 10-12 s bin has an RMSE of around 0.77 s and a bias of nearly zero. The mixed signal wave and turbulence signal in the profile's band of 8-10s provide the greatest difficulty for WaveConvLSTM in extracting the signal wave. There is likely to be the most room for improvement on this bin, as illustrated in \autoref{fig:WaveConvLSTM_ErrorStratification}.

\textbf{[Insert Fig. 6 here]}

\subsubsection{PtLSTM }
\begin{figure}[H]
    \centering
    \includegraphics[width=0.95\linewidth]{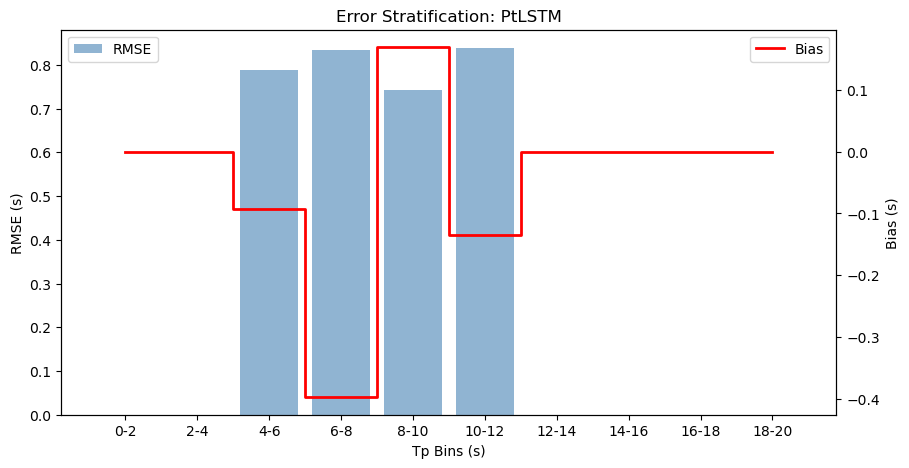}
    \caption{Shows the error stratification graph for PtLSTM model}
    \label{fig:PtLSTM_ErrorStratification}
\end{figure}

PtLSTM achieved a skill score of WS: 0.9747, and an SI: 0.0993. It shows an error stratification with a highest RMSE value in the range of 6-8 s and 10-12 s (around 0.83 s) and a lesser relative value in the 8-10 s range (around 0.75 s). However, it is interesting that the model is ‘under predicting' the data in the two bins at 4-6 s (~0.08 s) and 10-12 s (~0.10 s), i.e. the model is negative biased. PtLSTM has a WS of 0.9747 and a SI of 0.0993 (lowest of all skill scores) and its low SI verifies close error spread. In the case of high quality supervised data, the recurrent structure of PtLSTM can capture the temporal frequency patterns and can be a powerful system for tracking the overall temporal dynamics of the period of the waves as shown in \autoref{fig:PtLSTM_ErrorStratification}.

\textbf{[Insert Fig. 7 here]}

\subsubsection{PtAttnCNN (TinyWaveNets)}
\begin{figure}[H]
    \centering
    \includegraphics[width=0.95\linewidth]{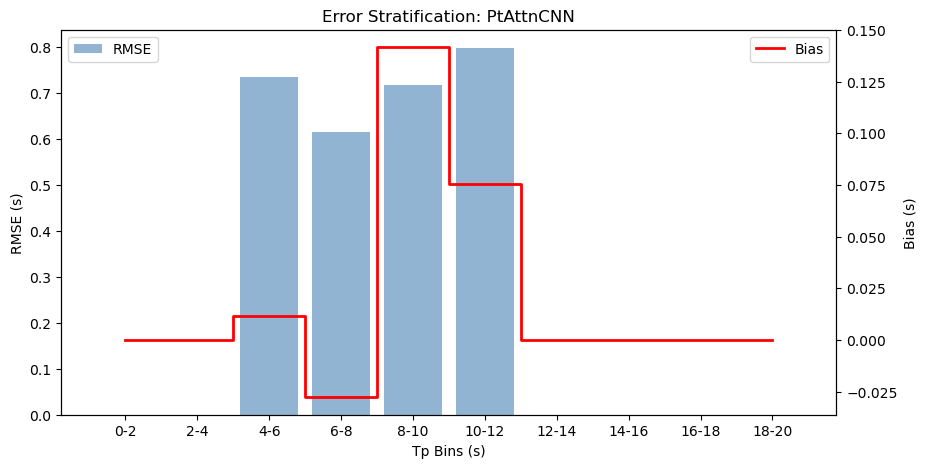}
    \caption{Shows the error stratification graph for PtAttnCNN model}
    \label{fig:PtAttnCNN_ErrorStratification}
\end{figure}

The skills of the other architectures were lower, with the best results being obtained by PtAttnCNN (TinyWaveNets), which achieved WS: 0.9767 and SI: 0.0924. This is also the backbone used for the Lambda ablation study where regularization in Physics is applied at $\lambda = 5.0$, to further reach WS: 0.9811 and SI: 0.0892, the overall best performance in this study. The stratification of the errors show that the maximum RMSE of the data are about 0.80 s contained within the 10-12 s bin, while the bin with 8-10 s data has high RMSE of data (~0.72 s) and the maximum positive bias is ~+0.15 s. The 6-8 s bin is the bin with the fewest errors and the model is optimal with the short to mid term period waves. The most restrictive collective prediction range and best trend, following performance is obtained with attention mechanism and CNN-GRU feature extraction, respectively, while for an operational monitoring system of coastal waves, where consistency of skill matters, the selected architecture is the combination of PtAttnCNN (TinyWaveNets) and CNN feature extraction, as shown in \autoref{fig:PtAttnCNN_ErrorStratification}.

\textbf{[Insert Fig. 8 here]}

\subsection{Ablation Analysis: The Role of Physics Weight ($\lambda$)}
The physics weight is sensitive, and has a characteristic recovery curve in the skill scores. While WS improves slightly from $\lambda = 0.0$ (0.9664) to $\lambda = 0.1$ (0.9707), it dips marginally at $\lambda = 1.0$ (0.9696) before recovering strongly at $\lambda = 5.0$ (0.9811). This changes in behaviour demonstrates that moderate regularisation of the physics (with physical weights) can have a short term adverse effect on the performance of data-driven fitting on weakly supervised labels, with a strong regularization on physics being the best.

\par
\begin{figure}[H]
    \centering
    \includegraphics[width=0.95\linewidth]{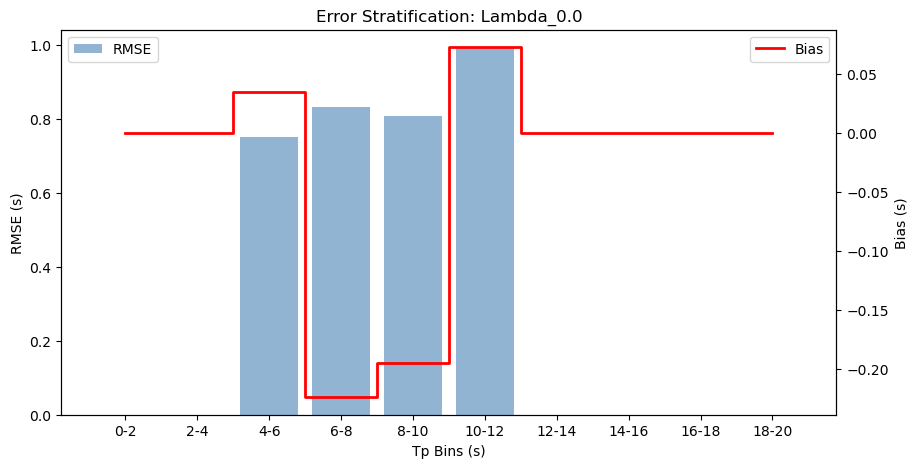}
    \caption{Error Stratification for Baseline ($\lambda = 0.0$)  – RMSE and Bias across $T_p$ Bins model}
    \label{fig:Lambda_0.0_ErrorStratification}
\end{figure}

\textbf{Baseline ($\lambda = 0.0$):} The (SI: $0.1099$, WS: $0.9664$) dThe stratification graph shows that the RMSE is maximum at about 10-12 s, and the bias is near zero (+0.05 s) at about 4-6 s, but is ~-0.20 s in the 6-8 s and 8-10 s bins, where there are no physical constraints, as shown in \autoref{fig:Lambda_0.0_ErrorStratification}.

\textbf{[Insert Fig. 9 here]}

\par
\begin{figure}[H]
    \centering
    \includegraphics[width=0.95\linewidth]{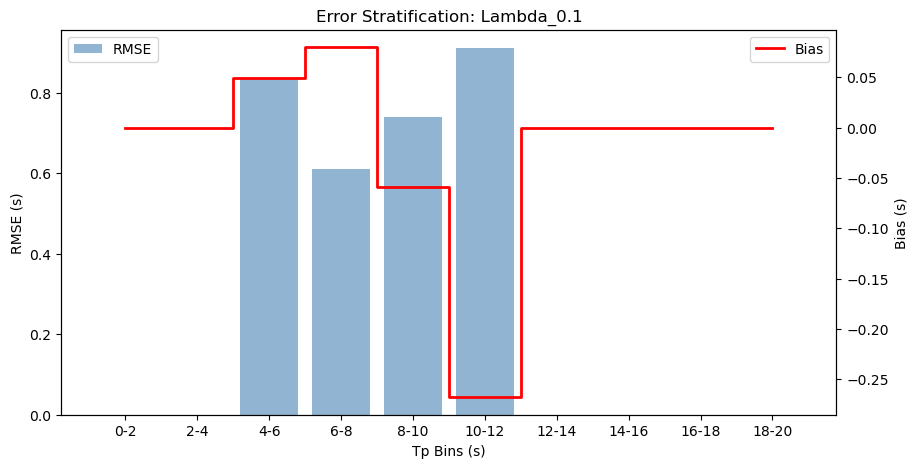}
    \caption{Error Stratification for Transition ($\lambda = 0.1$)  – RMSE and Bias across $T_p$ Bins model}
    \label{fig:Lambda_0.1_ErrorStratification}
\end{figure}

\textbf{Transition ($\lambda = 0.1$ to $1.0$):} Compared to the baseline, at $\lambda = 0.1$ skill goes up (WS: 0.9707, SI: 0.0972) because it is slightly regularized, which is used to directly constrain predictions. Only at $\lambda = 1.0$ (WS: 0.9696, SI: 0.1010) however, skill decreases a little below $\lambda = 0.1$, indicating that the moderate physics weighting momentarily out-competes the data-driven determination of the noisy training labels. The biases are greatest in the 8-10s bin and 10-12s bin with maximum biases of approximately +0.30s and the RMSE is greatest of approximately 0.88s in the two bins of stratification 8-10s and 10-12s. This matches the physics loss starting to draw predictions inwards towards the interior of the valid range before convergence is achieved, as illustrated in \autoref{fig:Lambda_0.1_ErrorStratification}.

\textbf{[Insert Fig. 10 here]}

\begin{figure}[H]
    \centering
    \includegraphics[width=0.95\linewidth]{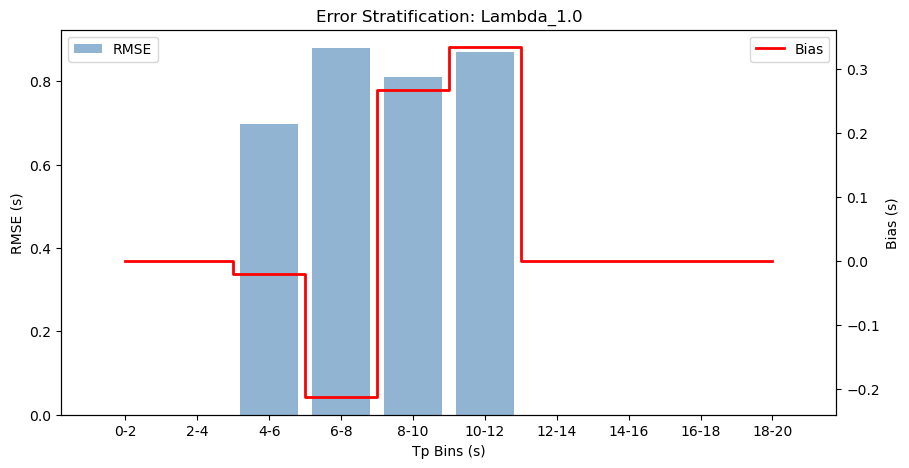}
    \caption{Error Stratification for Transition ($\lambda = 1.0$  – RMSE and Bias across $T_p$ Bins model}
    \label{fig:Lambda_1.0_ErrorStratification}
\end{figure}

The error stratification for the transition configuration at $\lambda = 1.0$ is shown in \autoref{fig:Lambda_1.0_ErrorStratification}.

\textbf{[Insert Fig. 11 here]}

\par
\begin{figure}[H]
    \centering
    \includegraphics[width=0.95\linewidth]{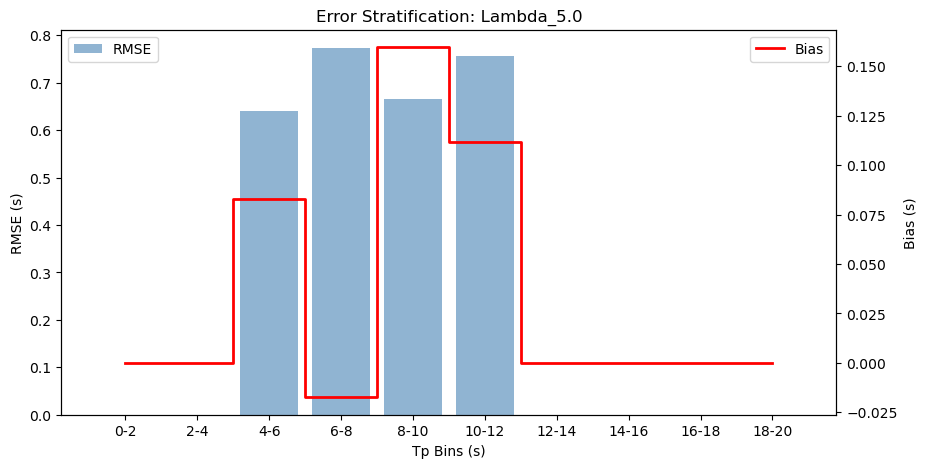}
    \caption{Error Stratification for Optimized Framework ($\lambda = 5.0$)  – RMSE and Bias across $T_p$ Bins model}
    \label{fig:Lambda_5.0_ErrorStratification}
\end{figure}

\textbf{Optimized Framework ($\lambda = 0.1$ to $5.0$):} The stratification error is most limited at $\lambda = 5.0$, where the maximum is at 0.77 s in the 6-8 s bin and the 10-12 s bin, and the bias is more or less the same as with the other settings with the bias being +0.09 s in each of the active bins. This holds true in good physics regularization that leads to a smaller increase in RMSE and a smaller bias drift throughout the entire range of wave periods. SI: 0.0892 and WS: 0.9811, representing the best overall configuration, as shown in \autoref{fig:Lambda_5.0_ErrorStratification}.

\textbf{[Insert Fig. 12 here]}

\subsection{Physical Guardrails and Distribution Alignment}
The distribution analysis and the ablation study agree that the Standard CNN (No Physics) often forecasts periods of less than 8 s, which is below the limit for the valid period: 

\begin{itemize}
 \setlength{\itemsep}{2pt}
 \setlength{\leftskip}{1em}
\item Leverages Pretrained Knowledge: The Synthetic Pretraining using Airy Linear Wave Theory provided a robust foundation that allowed the models to generalize from simulated deep-water waves to complex real-world nearshore environments.
\item Filters Spurious Signals: By utilizing Automated ROI Detection via temporal variance, the framework ignores static coastal features and focuses purely on the hydrodynamic intensity signal $s[t]$.
\item Respects Oceanographic Bounds: Strictly adheres to the frequency masking $0.05$--$0.5\,\mathrm{Hz}$ enforced during the FFT labeling and training phases.
\end{itemize}

\subsection{Computational Efficiency and Real-Time Feasibility}
The monitoring system of coastal monitoring stations requires a trade-off between forecast accuracy and computational cost, especially time and equipment. \hyperref[tab:model_complexity]{Table~\ref{tab:model_complexity}} summarizes the complexity of the system and the time required for the required parameters, showing the variation in the size of the measurement and the speed of understanding.

\textbf{[Insert Table 10 here]}

\begin{table}[h]
\caption{Model Complexity and Inference Latency Analysis}\label{tab:model_complexity}
\renewcommand{\arraystretch}{1.5}
\begin{tabular}{@{}llll@{}}
\toprule
\textbf{Model Architecture} & 
\textbf{Parameters (M)} & 
\textbf{Inference Latency (ms)} & 
\textbf{Operational Status} \\
\midrule

TinyWaveNet + $\lambda = 5.0$ (Proposed)
& 0.28 
& 0.53 
& Ultra-Lightweight \\

PtAttnCNN
& 0.28 
& 0.63 
& Real-Time Capable \\

LtViViT
& 2.36 
& 1.20 
& Optimal Balance \\

PtLSTM
& 12.52 
& 3.61 
& High Resource Demand \\

WaveConvLSTM
& 0.24 
& 57.01 
& Latency Bottleneck \\

\botrule
\end{tabular}
\end{table}

\FloatBarrier

\subsection{Physical Consistency and Model Auditing}
In order to verify that the physics-inspired TinyWaveNet model produces scientifically valid predictions, a post hoc interpretability audit of the physics-constrained PtAttnCNN (Lambda = 5.0) model was conducted at the clip-level across all 60-frame sliding windows from 13 Golden Set scenes. This resulted in a total of 6,926 clip-level audit evaluations. Scene-level statistics summarizing clip-level scores were subsequently calculated for each of the 13 scenes. This more granular clip-level analysis ensures that all temporal variability contained within each coastal scene can be accounted for, as opposed to evaluating just one representative frame.

\subsubsection{Spatial Focus via Grad-CAM Saliency}

As the PtAttnCNN (TinyWaveNet) architecture is convolutional instead of transformer, spatial interpretability could be evaluated based on spatial saliency provided by Grad-CAM instead of attention scores. Grad-CAM calculates gradients of the predicted output with respect to the input pixel intensities, generating a heatmap of image pixels that are responsible for $T_p$ predictions.

For the Physics-Guided TinyWaveNet model, the mean surf zone focus is 61.28\% across all 6,926 clip-level evaluations (over all 13 scenes).
As the surf zone focus metric reflects the percentage of time model attends to hydrodynamically active regions near shore, the high score of 61.28\% suggests that the majority of $T_p$ prediction attention focuses on hydrodynamically active surf zone rather than static coast features like land, sky, and infrastructure. The surf zone focus per scene varies from 57.4\% to 66.6\% due to the variability of the extent of wave activity zones in different Golden Set scenes.

\subsubsection{Physics-AI Spatial Overlap (IoU)}
In order to assess how well model’s spatial focus aligned with wave activity location defined through physics principles, a measure of spatial intersection over union (IoU) between Grad-CAM saliency map and physics-based wave crest position map is used. In particular, a physics-based wave crest location map was produced by detecting edges in the middle of sequence frame using Laplacian filter as a physics-derived heuristic for detecting wave crests. The Intersection over Union was then calculated between each Grad-CAM spatial saliency map and the physics-based wave crest position map, and then averaged across all 6,926 clip levels.
The average Physics-AI Intersection-over-Union (IoU) between all 6,926 clip evaluations was 0.103. Even though this IoU number is relatively low, it is a fair assessment of the inherent complexity involved in aligning a large-scale spatial saliency map to a sparse, binary crest mask, especially within dynamic surf zones with fast-shifting crest locations throughout the frames.

\subsubsection{Temporal Wave Theory Correlation}
To measure the temporal consistency of the AI prediction with regard to physically realistic wave motion, a Pearson correlation analysis was performed between the predicted $T_p$ values and the pixel-speed estimates calculated based on the temporal cross-correlation between the spatial transects, averaged across all 6,926 clip evaluations. In order to derive the pixel-speed, a cross-correlation analysis between the spatial intensity transects extracted from the first and the middle frames of each 60-frame clip was computed, providing an estimate of the wave propagation speed that matches the assumptions of linear wave theory, whereby the period and wave celerity are related by the dispersion relation.

The Pearson correlation coefficient R between the predicted $T_p$ values and estimated pixel speeds was 0.842, which is indicative of very high agreement between the predicted periods and the physical observations, thus supporting the claim that Physics-Guided TinyWaveNet has been able to learn the physical connection between wave period and the propagation speed.

\subsubsection{Synthesis}
In summary, three audit criteria provided converging support for the physical validity of the proposed framework. High surf zone focus (61.28\%) confirms the effectiveness of the automatic ROI detection in focusing model attention on the active region. The Physics-AI IoU (average 0.103) demonstrates that model attention exhibits consistent spatial alignment with respect to wave crest activity. The high Pearson correlation coefficient (R = 0.842) confirms that model predictions are consistent with physical motion patterns rather than artifacts present in the image data as presented in \hyperref[tab:physics_guided]{Table~\ref{tab:physics_guided}}.

\textbf{[Insert Table 11 here]}

\begin{table}[h]
\caption{Physics-Guided TinyWaveNet ($\lambda = 5.0$) Auditing Summary}\label{tab:physics_guided}
\renewcommand{\arraystretch}{1.5}
\begin{tabular}{@{}p{1.5cm}p{1.5cm}p{1cm}p{2cm}p{5cm}@{}}
\toprule
\textbf{Metric} & \textbf{Evaluation Level} & \textbf{Mean} & \textbf{Range (Min--Max)} & \textbf{Interpretation} \\
\midrule
Surf Zone Focus (\%) 
& Scene-level (aggregated from 6,926 clips) 
& 61.28\% 
& 57.4\% -- 66.6\% 
& Model directs majority of spatial attention toward the hydrodynamically active nearshore region across all 13 scenes \\

Physics-AI Overlap (IoU) 
& Clip-level (6,926 clips) 
& 0.103 
& 0.065 -- 0.111 
& Systematic spatial alignment between Grad-CAM saliency and Laplacian-detected wave crest locations; modest absolute value reflects sparse crest masks in dynamic surf zones \\

Wave Theory Correlation ($T_p$ vs.\ Pixel Speed) Pearson $R$
& Clip-level (6,926 clips) 
& 0.842 
& N/A 
& Strong agreement between predicted $T_p$ and physically derived pixel speed estimates, consistent with linear wave theory dispersion relationship \\
\botrule
\end{tabular}

\vspace{0.3cm}

\footnotesize{
\textit{Note:} Clip-level evaluation was performed across all 60-frame sliding windows extracted from the 13 Golden Set scenes (6,926 clips total). Scene-level Surf Zone Focus was obtained by averaging clip-level results within each scene.
}

\end{table}
\FloatBarrier

\section{Conclusion}
The findings of this study demonstrate the potential of low-cost coastal video imagery for physically consistent estimation of nearshore wave peak periods in operational monitoring environments. The framework includes three synergic components: (i) automated temporal-variance ROI detector which is independent of network backbone architecture, (ii) the three-stage Sim-to-Real Transfer learning, and (iii) the three-stage Sim-to-Real Mapping. These feature (i) pipeline for synthetic Airy wave pre-training, optical-flow Silver pre-training and Golden expert fine-tuning, and (ii) a hydrodynamic physics-informed loss function, which is suitable for the South Atlantic range of waves (2-20 s) ideal for deployment at the edge with sub-second inference latency.
Finally, this study illustrates the dependence of the models on the different forms of oceanographic skill ranking, suggesting the need to report both oceanographic skill scores (WS, SI) and point to point accuracy (RMSE) in unison and fully for a complete and meaningful evaluation. LtViViT achieved the strongest instantaneous prediction accuracy, suggesting that factorized spatiotemporal attention effectively captures localized nearshore wave dynamics. The overall performance (smallest prediction spread across all wave period bins) is best for the model PtAttnCNN (TinyWaveNet) without physics regularization (WS: 0.9767, SI: 0.0924) and with strong physics regularization at $\lambda = 5.0$ (WS: 0.9811, SI: 0.0892).

The physics regularized PtAttnCNN ($\lambda = 5.0$) is recommended for operational use due to the low number of parameters (0.28 M), the shortest inference time (latency) in real-time (0.53 ms) and the best oceanographic skill scores. The advantages of the ROI optimization were found to be the decrease of the RMSE by 27\%, and the absence of physically unrealistic predictions and the spurious spectral peaks. The temporal smoothness analysis also showed that the $T_p$ estimation is consistent for the allowed number of consecutive frames in the video for the various oceanographic conditions (foam coverage and illumination).
The framework has additional practically relevant attributes in addition to being correct. A good physics-informed warm start is demonstrated by a 86.93\% reduction in the validation RMSE from 2.4012s at Epoch 1 to 0.3138s at Epoch 25 as the synthetic pre-training was used for Airy wave sequences. The moderate increase, but faster one was seen in the first two epochs (60.3\%) signifying that there is a fast convergence at the initial stages of the training process. The lightweight LtViViT variant (2.36 M parameters) can be installed at the edge and the ROI detection module is uncoupled and does not need to be adjusted according to the camera angle.
The characteristics of the proposed framework are well suited and cost-effective to be applied to any environment that has high data volume and/or complex logistics in the nearshore area as an alternative to the conventional in-situ sensor arrays and complex deployment of a radar system. The 13 scenarios being validated now cover some open-source videos from the sources mentioned in section 2.1.1 and also from South Atlantic coast of Namibia in a spatially representative area with a consistent long-period swell climate, in open beach, rocky headland and rocky breakwater areas. Future work will cover other world wave climates, e.g., wind-sea dominated (North Sea, Mediterranean) and other mixed swell regimes (Pacific), will be further validated and complementary parameters (significant wave height (Hs), wave direction, runup) will be included as will be hybrid transformer-recurrent architectures that could offer further latency reduction. 

In the end, it is an economically viable physically robust monitoring device for long-term coastal monitoring and it can be combined with passive video streams to provide support to the engineering, research and coastal management communities for erosion and accretion control, beach morphology prediction, and adaptation to climate change.

\section*{Data Availability Statement}

The datasets generated and/or analyzed during the current study are not publicly available due to data management considerations but are available from the corresponding author upon reasonable request.

\section*{Statements and Declarations}
\subsection*{Funding:} This research was supported by the German Academic Exchange Service (Deutscher Akademischer Austauschdienst, DAAD), through its In-Country/In-Region Scholarship Programme, which contributed to the doctoral research activities of the corresponding author, and by computational resources, including high-performance computing facilities, provided by the Namibia University of Science and Technology. No funding body had any role in the study design, data collection and analysis, decision to publish, or preparation of the manuscript.
\subsection*{Competing Interests:} The authors declare that they have no known competing financial interests or personal relationships that could have appeared to influence the work reported in this paper.
\subsection*{Author Contributions:} Abubakar Hamisu Kamagata: conceptualization, methodology, software, formal analysis, investigation, data curation, writing — original draft, visualization. Dharm Jat Singh: supervision, conceptualization, methodology, writing — review and editing. Attlee Munyaradzi Gamundani: supervision, methodology, writing — review and editing. Abhishek Srivastava: methodology, validation, writing — review and editing. Paramasivam Saravanakumar: resources, domain validation, writing — review and editing. All authors read and approved the final manuscript.
\subsection*{Ethics Approval:} This study did not involve human participants, human data, or animals, and did not require ethics committee approval. All video data used were either sourced from open-access repositories and commercial stock-video libraries under applicable licensing terms, or newly recorded by the research team at publicly accessible coastal sites.
\subsection*{Consent to Participate:} Not applicable.
\subsection*{Consent for Publication:} Not applicable.
\subsection*{Data Availability:} The datasets generated and/or analyzed during the current study are not publicly available due to data management considerations but are available from the corresponding author upon reasonable request.
\subsection*{Code Availability:} The custom code used for ROI detection, synthetic pretraining, model training, and evaluation is available from the corresponding author upon reasonable request.








\bibliography{sn-bibliography}

\end{document}